\documentclass{article}

\usepackage{arxiv}

\usepackage[utf8]{inputenc} % allow utf-8 input
\usepackage[T1]{fontenc}    % use 8-bit T1 fonts
\usepackage{hyperref}       % hyperlinks
\usepackage{url}            % simple URL typesetting
\usepackage{booktabs}       % professional-quality tables
\usepackage{amsfonts}       % blackboard math symbols
\usepackage{nicefrac}       % compact symbols for 1/2, etc.
\usepackage{microtype}      % microtypography
\usepackage{lipsum}
\usepackage{graphicx}
\graphicspath{ {./images/} }

\usepackage{amsmath} % assumes amsmath package installed
\usepackage{amssymb}  % assumes amsmath package installed
\usepackage{booktabs}
\usepackage{siunitx}
\usepackage{multirow}
\usepackage[dvipsnames]{xcolor}
\usepackage{hyperref}

\title{Online Learning of Wheel Odometry Correction for Mobile Robots with Attention-based Neural Network}

\author{Alessandro Navone$^{1}$, Mauro Martini$^{1}$, Simone Angarano$^{1}$ and Marcello Chiaberge$^{1}$ %<-this % stops a space
\thanks{$^{1}$ Department of Electronics and Telecommunications, Politecnico di Torino, 10129, Torino, Italy. \tt\footnotesize \{firstname.lastname\}@polito.it}}
\author{
 Alessandro Navone \\
  Department of Electronics and Telecommunications \\
  Politecnico di Torino\\
  Torino, TO, 10129 \\
  \texttt{alessandro.navone@polito.it} \\
   \And
 Mauro Martini \\
  Department of Electronics and Telecommunications \\
  Politecnico di Torino\\
  Torino, TO, 10129 \\
  \texttt{mauro.martini@polito.it} \\
  \And
 Simone Angarano \\
  Department of Electronics and Telecommunications \\
  Politecnico di Torino\\
  Torino, TO, 10129 \\
  \texttt{simone.angarano@polito.it} \\
  \And
 Marcello Chiaberge \\
  Department of Electronics and Telecommunications \\
  Politecnico di Torino\\
  Torino, TO, 10129 \\
  \texttt{marcello.chiaberge@polito.it} \\
}

\begin{document}
\maketitle
\begin{abstract}
Modern robotic platforms need a reliable localization system to operate daily beside humans. Simple pose estimation algorithms based on filtered wheel and inertial odometry often fail in the presence of abrupt kinematic changes and wheel slips. Moreover, despite the recent success of visual odometry, service and assistive robotic tasks often present challenging environmental conditions where visual-based solutions fail due to poor lighting or repetitive feature patterns. In this work, we propose an innovative online learning approach for wheel odometry correction, paving the way for a robust multi-source localization system. An efficient attention-based neural network architecture has been studied to combine precise performances with real-time inference. The proposed solution shows remarkable results compared to a standard neural network and filter-based odometry correction algorithms. Nonetheless, the online learning paradigm avoids the time-consuming data collection procedure and can be adopted on a generic robotic platform on-the-fly.
\end{abstract}

% keywords can be removed

\section{Introduction}
\label{sec:intro}
\begin{figure}
    \centering
    \includegraphics[trim={0 0 6cm 0},clip,width=\columnwidth]{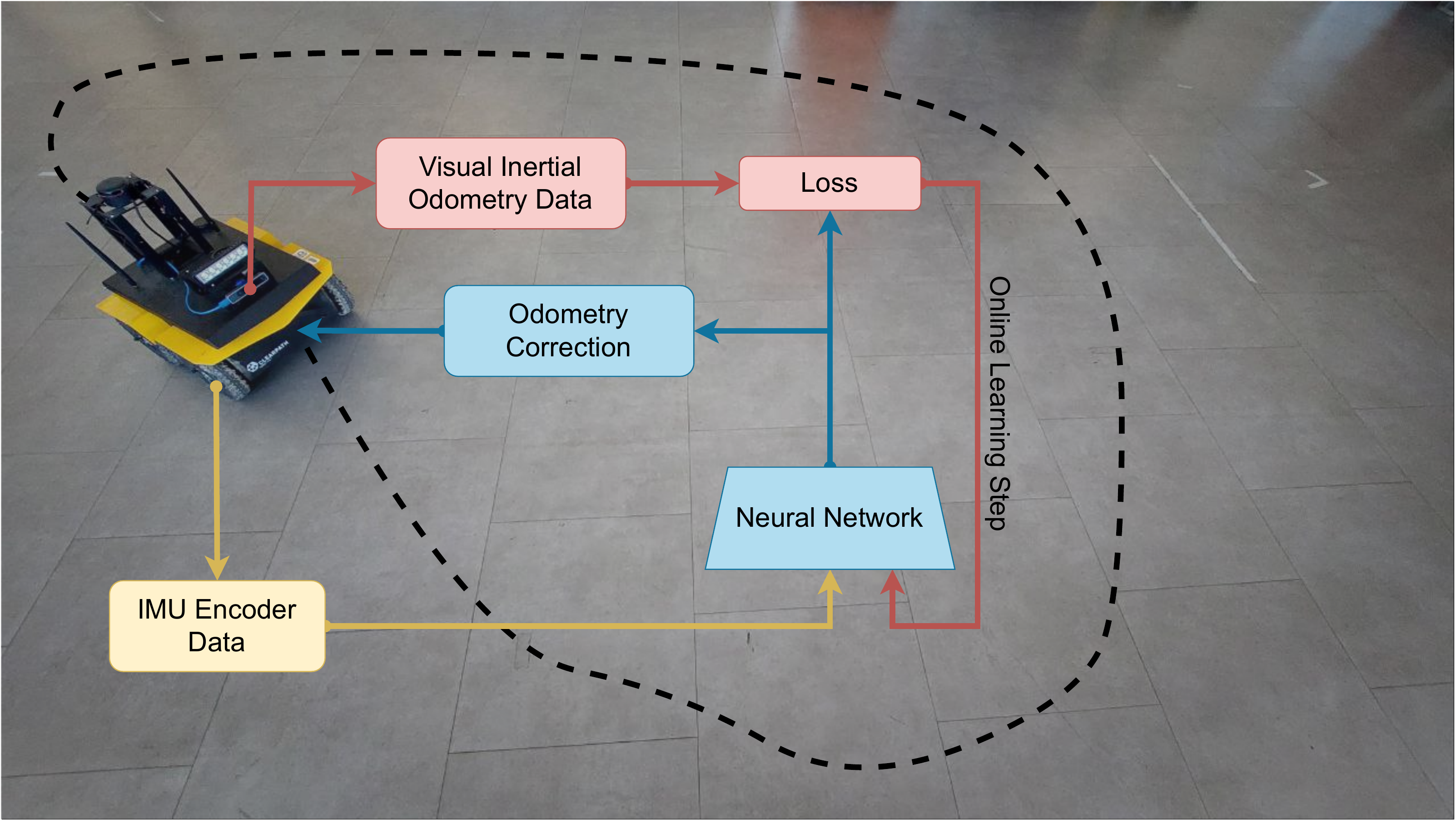}
    \caption{Diagram of the proposed approach. \textcolor{BrickRed}{Red} blocks and arrows refer to the online training phase, \textcolor{RoyalBlue}{blue} ones to the model inference stage, and \textcolor{Dandelion}{yellow} ones to the odometric input data.}
    \label{fig:schema}
\end{figure}

Wheel odometry (WO) and inertial odometry (IO) are the simplest forms of self-localization for wheeled mobile robots \cite{mohamed_survey_2019}. However, extended trajectories without re-localization, together with abrupt kinematic and ground changes, drastically reduce the reliability of wheel encoders as the unique odometric source. For this reason, visual odometry (VO) has recently emerged as a more general solution for robot localization \cite{wang2020approaches}, relying only on the visual features extracted from images. Nonetheless, service and assistive robotics platforms may often encounter working conditions that forbid the usage of visual data. Concrete scenarios are often related to the lack of light in indoor environments where GPS signals are denied, as occurs in tunnels exploration \cite{tardioli2019ground, seco2022robot} or in assistive nightly routines \cite{eirale2022marvin, eirale2022human, tamantini2021robotic}. Repetitive feature patterns in the scene can also hinder the precision of VO algorithms, a condition that always exists while navigating through empty corridors \cite{gupta2020corridor} or row-based crops \cite{martini2022}. Therefore, an alternative or secondary localization system besides VO can provide a substantial advantage for the robustness of mobile robot navigation. Wheel-inertial odometry is still widely considered a simple but effective option for localization in naive indoor scenarios. However, improving its precision in time would extend its usage to more complex scenarios. Previous works tackle the problem with filters or simple neural networks, as discussed in Section\ref{subsec:related}. Learning-based solutions demonstrate to mitigate the odometric error at the cost of a time-consuming data collection and labeling process. Recently, online learning has emerged as a competitive paradigm to efficiently train neural networks on-the-fly avoiding dataset collection \cite{perez-sanchez_review_2018}. In this context, this work aims at paving the way for a learning-based system directly integrated into the robot and enabling a seamless transition between multiple odometry sources to increase the reliability of mobile robot localization in disparate conditions. Figure \ref{fig:schema} summarizes the proposed methodology schematically.

\subsection{Related Works}
\label{subsec:related}
%Wheel odometry
Several studies have explored using machine learning techniques to estimate wheel odometry (WO) in mobile robotics applications. Approaches include different feed-forward neural networks (FFNN) \cite{xu_estimating_2009}, of which, in some cases, the output has been fused with other sensor data \cite{li_neural_2020}, and long short-memory (LSTM) NN, which have been applied to car datasets \cite{onyekpe_learning_2020}. These approaches show a promising improvement in WO accuracy, which is crucial for mobile robotics applications.

%%%%%%%%%%%%%%%%%%%%%%%%%%%%%%%%%%%%%%%%%%%%%%%%%%%%%%%%%%%%%%%%%%%%%%%%%%%%%%%%%%%%%%%%%%%%%%
\begin{figure*}[ht]
    \centering
    \includegraphics[width=0.9\textwidth]{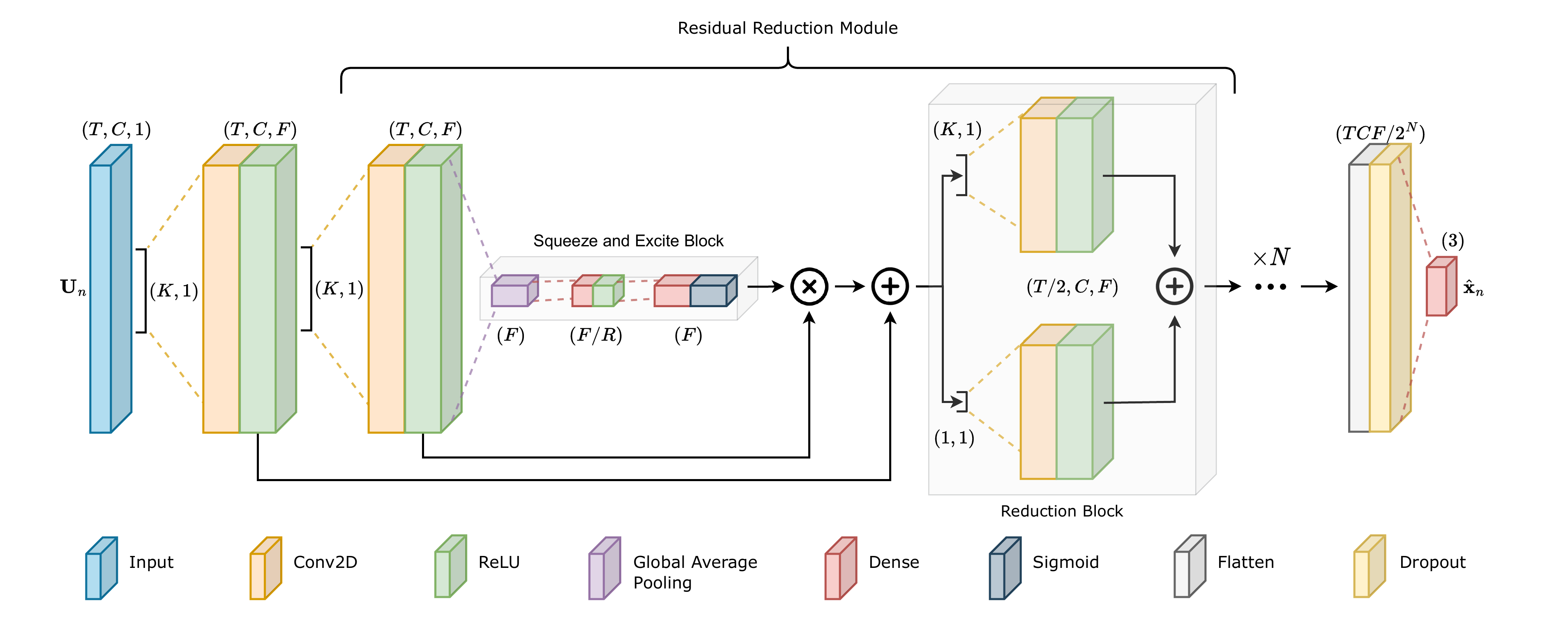}
    \caption{Architecture of the proposed model. The batch dimension is omitted for better clarity.}
    \label{fig:architecture}
\end{figure*}

%imu
Many works have focused on using Inertial Measurement Unit (IMU) data in mobile robots or other applications, such as person tracking using IMU data from cell phones \cite{chen_deep_2021}. One system was improved by implementing a zero-velocity detection with Gate Recurrent Units (GRU) neural network \cite{chen_improved_2020}. Another study used an Extended Kalman Filter (EKF) to estimate positions and velocities in real-time in a computationally lightweight manner \cite{solin_inertial_2018}. Additionally, a custom deep Recurrent Neural Network (RNN) model, IONet, was used to estimate changes in position and orientation in independent time windows \cite{chen_ionet_2018}. Some studies used a Kalman Filter (KF) to eliminate noise from the accelerometer, gyroscope, and magnetometer sensor signals and integrate the filtered signal to reconstruct the trajectory \cite{botero_valencia_simple_2017}. Another KF approach has been combined with a Neural Network to estimate the noise parameters of the filter \cite{brossard_ai-imu_2020}. 

Several neural network architectures have been proposed to predict or correct IO odometry over time. For example, a three-channel LSTM was fed with IMU measurements to output variations in position and orientation and tested on a vehicle dataset \cite{abolfazli_esfahani_aboldeepio_2020}. Another LSTM-based architecture mimics a kinematic model, predicting orientation and velocity given IMU input data. Studies have investigated the role of hyper-parameters in IO estimation \cite{dugne-hennequin_understanding_2021}.

%IMU e WO
Sensor fusion of wheel encoder and IMU data is a common method for obtaining a robust solution. One approach involves fusing the data with a Kalman Filter, which can assign a weight to each input based on its accuracy \cite{jinglin_shen_localization_2011}. A Fully Connected Layer with a convolutional layer has been employed for estimating changes in position and orientation in a 2D space over time in an Ackermann vehicle, along with a data enhancement technique to improve learning efficiency \cite{zhang_learning_2021}. Additionally, a GRU RNN-based method has been proposed to compensate for drift in mechanum wheel mobile robots, with an in-depth fine-tuning of hyper-parameters to improve performance \cite{canbek_drift_2022}.

\subsection{Contributions}\label{subsec:contribution}
%contribution
In this work, we tackle the problem of improving wheel-inertial odometry by learning how to correct it online with an efficient artificial neural network. At this first stage, the study has been conceived to provide the robot with a more reliable, secondary odometric source in standard indoor environments where the working conditions for VO can temporarily vanish, as in the case of robots for domestic night surveillance or assistance.
The main contribution of this work can be summarized as:
\begin{itemize}
    \item A novel online learning approach for wheel-inertial odometry correction which allows avoiding complex trajectory data collection and can be directly included in a ROS 2 system;
    \item An efficient model architecture to preserve both easy online training and fast inference performance.
\end{itemize}
Nonetheless, a validation dataset of sensor data has been collected with the robot following different trajectories to conduct extensive experiments and comparisons with state-of-the-art offline methods.

\section{Methodology}
\label{sec:methodology}
\subsection{Problem Formulation}\label{subsec:problem}
The position of a robot at time $t$ referred to the starting reference frame $\textbf{R}_{0}$ can be calculated by accumulating its increments during time segments $\delta t$. The time stamp $n$ refers to the generic time instant $t = n\delta t$. The state of the robot $\textbf{x}_n$ is defined by the position and orientation of the robot, such as:
\begin{equation}
    \textbf{x}_n = (x_n, y_n, \theta_n)^T,
\end{equation}
where $(x_n, y_n)$ is the robot's position in the 2D space and $\theta_n$ is its heading angle. 
Given the state, it is possible to parametrize the roto-translation $\textbf{T}^{m}_{0}$ matrix from the robot’s frame $\textbf{R}_{m}$ to the global frame $\textbf{R}_{0}$. Its first two columns represent the axes of the robot frame, and the last one is its position with respect to the origin.

The robot employed to develop this work is equipped with an IMU, which includes a gyroscope and an accelerometer, and two wheel encoders. Therefore, $\textbf{u}_n$ is defined as the measurement array referred to instant $n$, i.e.:
\begin{equation}
    \textbf{u}_n = 
    \begin{pmatrix}
    v_l, v_r, \ddot{x}, \ddot{y}, \ddot{z}, \dot{\theta}_x, \dot{\theta_y}, \dot{\theta}_z
    \end{pmatrix}^T,
\end{equation}
where $(v_l, v_r)$ are the wheels’ velocities, $(\ddot{x}, \ddot{y}, \ddot{z})$ are the linear accelerations and $(\dot{\theta}_x, \dot{\theta}_y, \dot{\theta}_z)$ are the angular velocities. The input $\textbf{U}_n$ to the proposed model consists in the concatenation of the last $N$ samples of the measurements $\textbf{U}_n = (\textbf{u}_{(n)}, \textbf{u}_{(n-1)}, \dots, \textbf{u}_{(n-N)})^T $. 
At each time sample, the state is updated as a function of the measurements $f(\textbf{U}_n)$: first, the change of the pose $\delta \hat{x} = f(\textbf{U}_n)$ of the robot is estimated, relative to the previous pose $\hat{\textbf{x}}_{n-1}$. Then, the updated state is calculated, given the transformation matrix obtained before, as:
\begin{equation}
    \hat{\textbf{x}}_n = \hat{\textbf{x}}_{n-1} \boxplus f(\textbf{U}_n) = \textbf{T}^m_{0(n-1)}\delta\hat{\textbf{x}}_n,
\end{equation}
where the operator $\boxplus$ symbolizes the state update.

\subsection{Neural Network Architecture}\label{subsec:NNarchi} % SIMO
As formalized in the previous section, the prediction of $\hat{\textbf{x}}_n \in \mathbb{R}^{3}$ from $\textbf{U}_n \in \mathbb{R}^{T\times C}$ is framed as a regression problem. The architecture we propose to solve this task is inspired to REMNet \cite{angarano_robust_2021,angarano2022ultra}, though it uses 2D convolutions instead of the original 1D convolutional blocks (Figure \ref{fig:architecture}). This modification aims at exploiting temporal correlations without compressing the channel dimension throughout the backbone. In particular, we keep the channel dimension $C$ separated from the filter dimension $F$. In this way, the first convolutional step with kernel $(K,1)$ and $F$ filters outputs a low-level feature map $f_1\in \mathbb{R}^{T\times C\times F}$. Then, a stack of $N$ Residual Reduction Modules (RRM) extracts high-level features while reducing the temporal dimension $T$. Each RRM consists of a residual ($Res$) block followed by a reduction ($Red$) module:
\begin{equation}
    RRM(x) = Red(Res(x))
\end{equation}
The $Res$ block comprises a 2D convolution with kernel $K\times 1$ followed by a Squeeze-and-Excitation (SE) block \cite{hu2018squeeze} on the residual branch. The SE block applies attention to the channel dimension of the features with a scaling factor learned from the features themselves. First, the block applies average pooling to dimensions $T$ and $C$. Then, it reduces the channel dimensionality with a bottleneck dense layer of $F/R$ units. Finally, another dense layer restores the original dimension and outputs the attention weights. After multiplying the attention mask for the features, the result is used as a residual and added to the input of the residual block. The $Red$ block halves the temporal dimension by summing two parallel convolutional branches with a stride of 2. The layers have kernels $K\times1$ and $1\times1$, respectively, to extract features at different scales. After $N$ RRM blocks, we obtain the feature tensor $f\in\mathbb{R}^{T\times C\times F/2^N}$, which is flattened to predict the output through the last dense layer. We also include a dropout layer to discourage overfitting. 

\subsection{Training Procedure}\label{subsec:training} 
The goal of this work consists of learning the positioning error of the robot using wheel odometry. Nonetheless, it is important to remark that, nowadays, visual-inertial odometry (VIO) is a standard approach on robotic platforms. This work does not aim to propose a more precise localization system but to learn wheel-inertial odometry as a second reliable localization algorithm available whenever visual approaches fail. 
 
We exploit a basic VIO system on the robot for the only training process since it enables a competitive online learning paradigm to train the model directly on the robot. Batch learning, the most used training paradigm, requires all the data to be available in advance. As long as the data are collected over time, the proposed method consists in training the network in a continuous way when a batch of N data is available. This approach has been tested extensively in \cite{chen_fine-tuning_2017}, demonstrating a negligible loss in accuracy compared to the batch-learning paradigm.

The proposed model's training consists of two main steps, which are repeated as long as new data are available. First, a batch of $N$ elements is collected, respectively, the input of the network $\textbf{U}_n$ and the expected output $\delta x$. Then, an update step is carried out using an SGD-based optimizer algorithm adopting a Mean Absolute Error loss function, which does not emphasize the outliers or the excessive noise in the training data.

%%%%%%%%%%%%%%%%%%%%%%%%%%%%%%%%%%%%%%%%%%%%%%%%%%%%%%%%%%%%%%%%%%%%%%%%%%%%%%%%

\begin{figure}[t]
    \centering
    \includegraphics[width=0.55\textwidth]{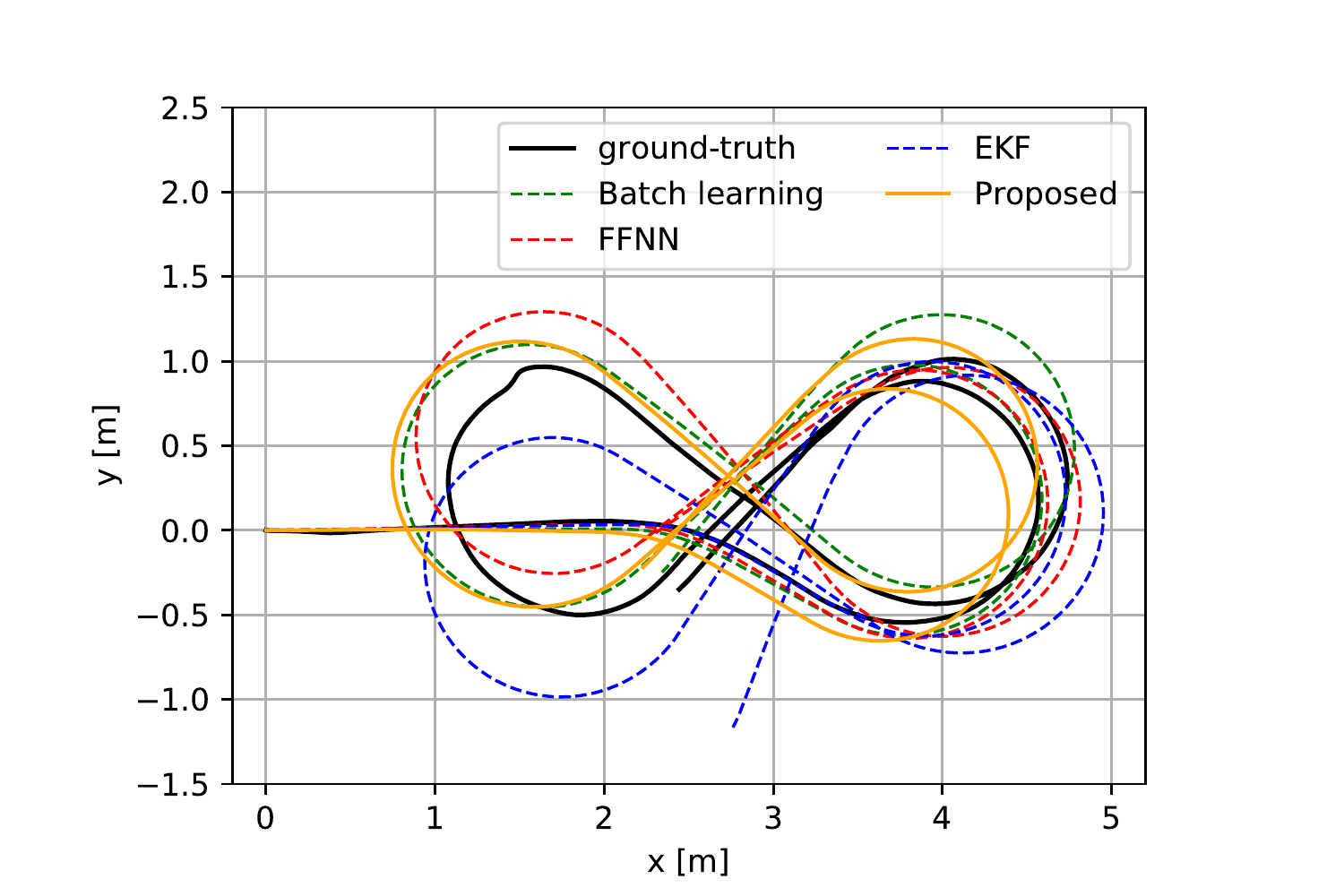}
    \caption{Infinite-shaped trajectories estimated by different methods. The data are collected during a total navigation time of about $60 s$.}
    \label{fig:fig_xy}
\end{figure}

%\begin{figure}[t]
%  \centering
%  \includegraphics[width=0.7\columnwidth]{abs_pos_err_1.pdf}
%  \includegraphics[width=0.7\columnwidth]{abs_yaw_err_1.pdf}
%\caption{Absolute error of position and orientation of different methods during the test performed on a subset of infinite-shaped trajectories. The considered subset is the same as figure \ref{fig:fig_xy}.}
%\label{fig:errors}
%\end{figure}

\begin{figure}[!tbp]
    \centering
    \begin{minipage}[b]{0.49\textwidth}
    \includegraphics[width=\textwidth]{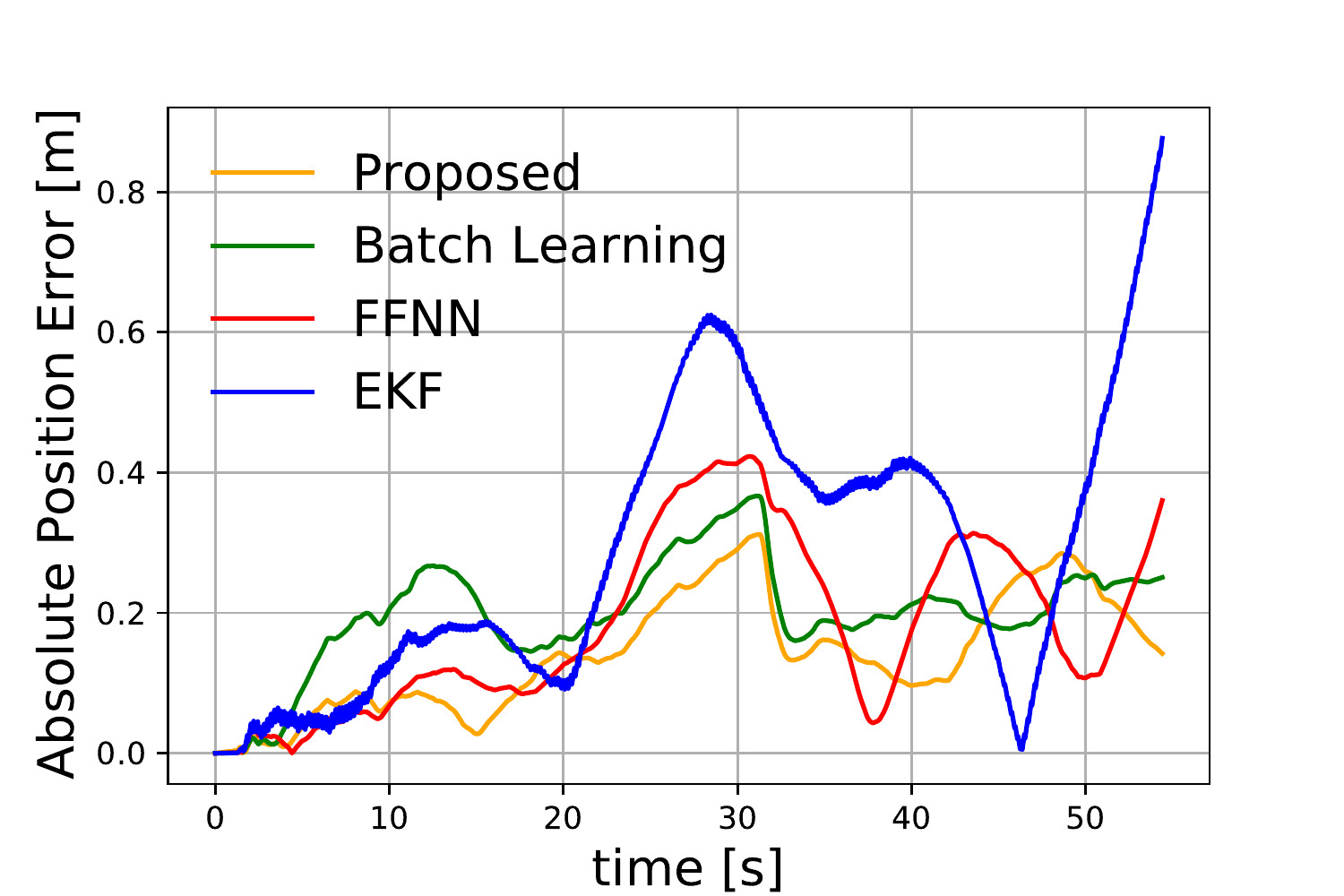}
    \end{minipage}
    \hfill
    \begin{minipage}[b]{0.49\textwidth}
    \includegraphics[width=\textwidth]{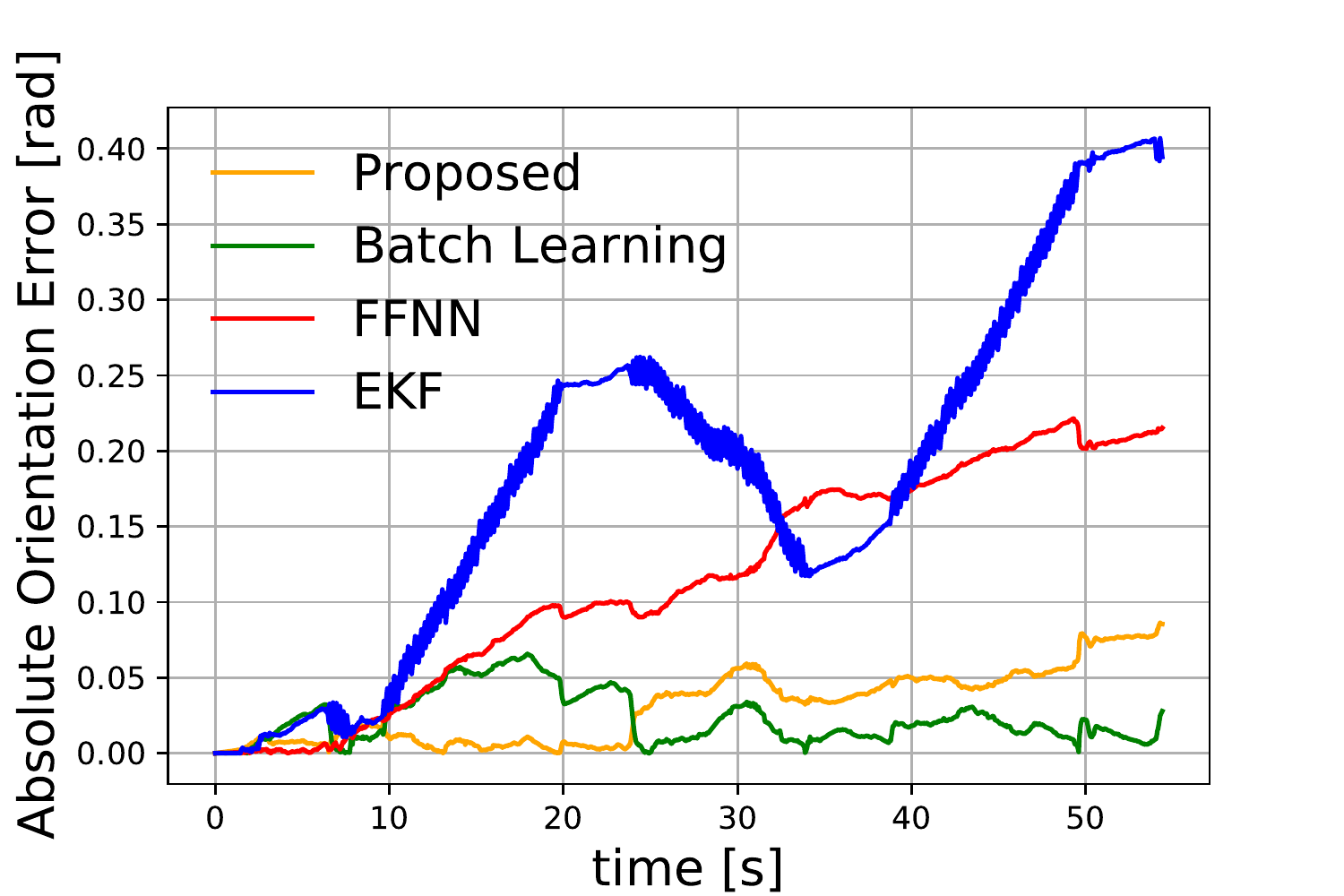}
    \end{minipage}
    \caption{Absolute error of position and orientation of different methods during the test performed on a subset of infinite-shaped trajectories. The considered subset is the same as figure \ref{fig:fig_xy}.}
    \label{fig:errors}
\end{figure}

%%%%%%%%%%%%%%%%%%%%%%%%%%%%%%%%%%%%%%%%%%%%%%%%%%%%%%%%%%%%%%%%%%%%%%%%%%%%%%%%
\section{Tests and Results}\label{sec:results}
In this section, the proposed approach is tested through extensive experimental evaluations. The model presented in Section \ref{subsec:NNarchi} has been trained with an incremental learning method and a classical batch training approach. Results obtained with a simple FFNN model and a standard localization solution based on an EKF are also discussed in the comparison. For this sake, both training processes have been accomplished on the same dataset, and all the tests have been executed on the same test set.

\subsection{Experimental Setting}
\label{subsec:paragrafo_exp}
The dataset used for the experiments was collected in a generic indoor environment. The employed robotic platform was a Clearpath Jackal\footnote{\url{https://clearpathrobotics.com/jackal-small-unmanned-ground-vehicle/}}, a skid-steer driving four-wheeled robot designed for indoor and outdoor applications.
All the code was developed in a ROS 2 framework and is tested on Ubuntu 20.04 LTS using the ROS 2 Foxy distro.

Since an indoor environment was considered, the linear velocity of the robot was limited to $0.4 m/s$ and its angular velocity to $1 rad/s$. The data from the embedded IMU and wheel encoders were used as inputs to the model. According to these assumptions, we used the robot pose provided by an Intel Realsense T265 tracking camera as ground truth. As the testing environment is a single room, the precision of the tracking camera is guaranteed to provide a drift of less than $1 \%$ in a closed loop path\footnote{\url{https://www.intelrealsense.com/tracking-camera-t265/}}. All the data have been sampled at $1/ \delta t = 25 Hz$.

The data were collected by teleoperating the robot around the room and recording the sensor measurements. For the training dataset, the robot has been moved along random trajectories. For the test dataset, critical situations when the skid-steer drive robot’s odometry is known to lose the most accuracy were reproduced, such as tight curves, hard brakings, strong accelerations, and turns around itself. The obtained training dataset consists of 156579 samples; 80\% have been used for training and 20\% for validation and hyperparameter tuning. The test dataset consists of 61456 samples.

The model hyperparameters have been tuned by performing a grid search using a batch learning process, considering a trade-off between accuracy and efficiency. In the identified model, we adopted $F=64$ filters, $N=2$ reduction modules, and a ratio factor $R=4$. Kernel size $K=3$ is used for all the convolutional layers, including the backbone. The input dimensions were fixed to $T=10$ and $C=8$. The former corresponds to the number of temporal steps, and it has been observed how a higher value appears to be superfluous. In contrast, a lower value leads to performance degradation. The latter value, $C$, corresponds to the number of input features, i.e., sensor measurements as described in \ref{sec:methodology}. 

We adopted Adam\cite{kingma_adam_2017} as the optimizer for the training. The exponential decay rate for the first-moment estimates is fixed to $\beta_1 = 0.9$, and the decay rate for the second-moment estimates is fixed to $\beta_2 = 0.999$. The epsilon factor for numerical stability is fixed to $\epsilon = 10^{-8}$. The optimal learning rate $\eta$ was experimentally determined as \num{1e-4} for batch learning. Conversely, the incremental learning process showed how a value of $\eta = \num{7e-5}$ avoided overfitting since the data were not shuffled. In both learning processes, a batch size of $B=32$ was used.

\begin{figure}[!tbp]
    \centering
    \begin{minipage}[b]{0.49\textwidth}
    \includegraphics[width=\textwidth]{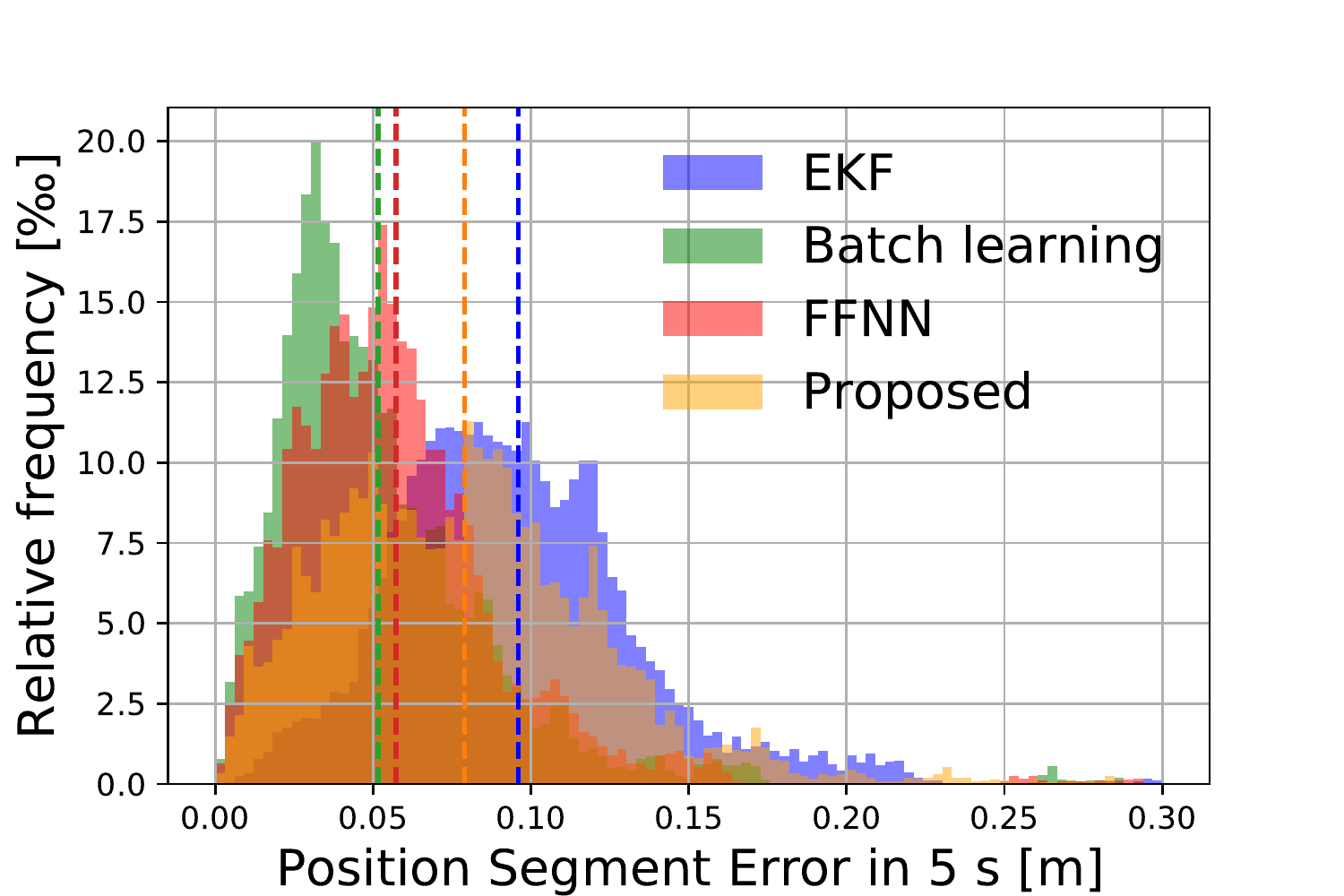}
    \end{minipage}
    \hfill
    \begin{minipage}[b]{0.49\textwidth}
    \includegraphics[width=\textwidth]{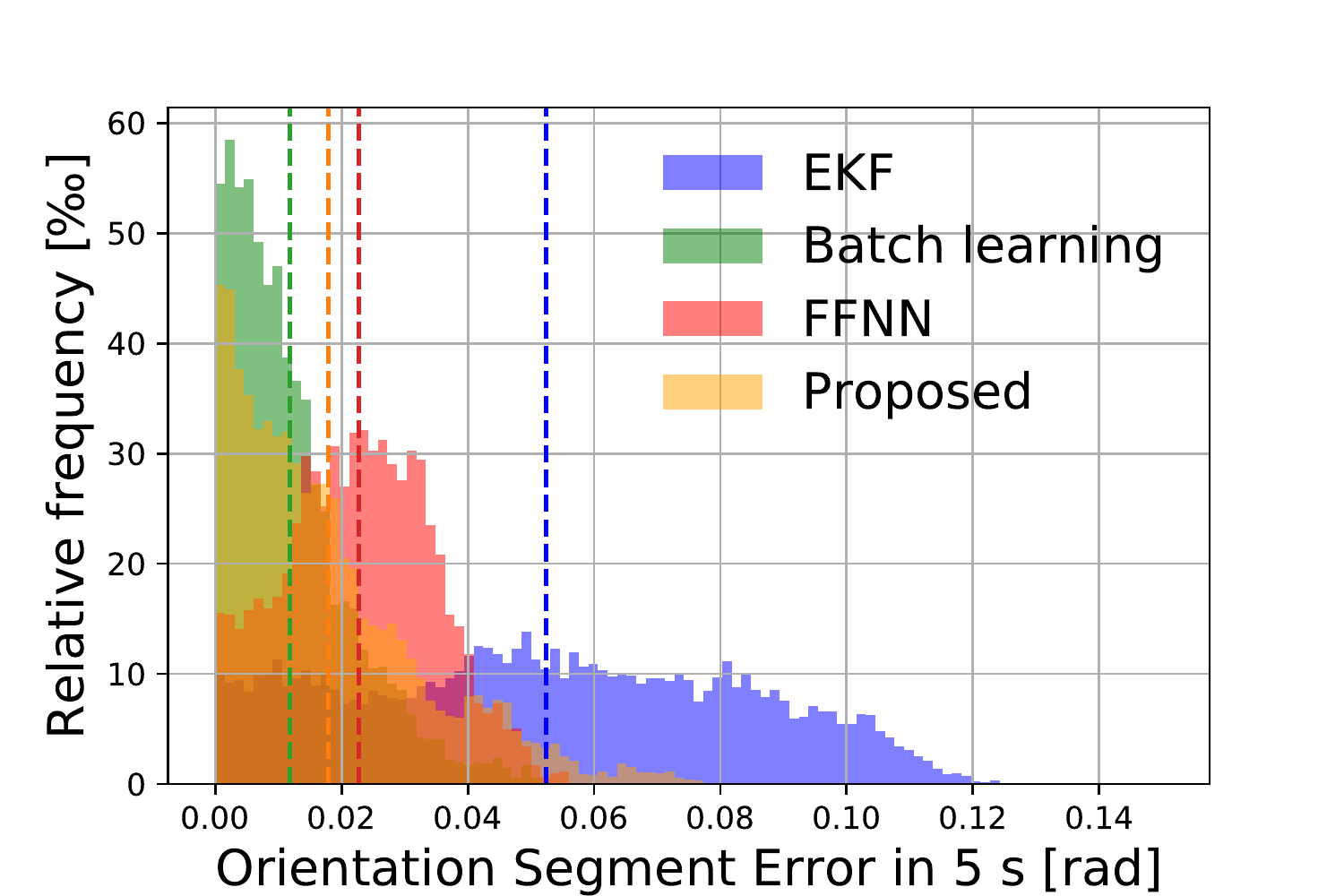}
    \end{minipage}
    \caption{Histograms of the SE error in position and orientation in section B of the test set.}
    \label{fig:histograms}
\end{figure}

%\begin{figure}[t]
%  \centering
%  \includegraphics[width=0.7\columnwidth]{hist_pos_SE_round.pdf}
%  \includegraphics[width=0.7\columnwidth]{hist_theta_SE_round.pdf}
%\caption{Histograms of the SE error in position and orientation in section B of the test set.}
%\label{fig:histograms}
%\end{figure}
%%%%%%%%%%%%%%%%%%%%%%%%%%%%%%%%%%%%%%%%%%%%%%%%%%%%%%%%%%%%%%%%%%%%%%%%%%%%%%%%
\subsection{Evaluation Metrics}\label{subsec:metrics}
To evaluate the performance of the proposed model, two different metrics were used \cite{peretroukhin_dpc-net_2018}:

\begin{table*}[t]
    \centering
    \caption{Performance comparison on the different test scenarios and the overall test set with the respective standard deviation. When trained with batch learning, the proposed architecture performs better than the FFNN proposed in \cite{zhang_learning_2021}. If trained online, it outperforms the common EKF-based localization method and achieves the results of the model trained offline.}
    \begin{tabular}{c c c c c c c}
    \toprule
    Test  &  Duration $[s]$ & Method & $m-ATE_{(x, y)} [m]$ & $m-ATE_{\theta} [rad]$ & $SE_{(x,y)} [m]$ &  $SE_{\theta} [rad]$   \\
    \midrule
    \multirow{4}{*}{A} & \multirow{4}{*}{998} & EKF & $0.692 \pm 0.213$ & $0.821 \pm 0.334$ & $0.099 \pm 0.043$ & $0.069 \pm 0.032$\\
    
     & & \textbf{Online Learning} & $0.292 \pm 0.098 $& $0.118 \pm 0.079$ & $0.071 \pm 0.038$ & $0.020 \pm 0.013$\\
    \cmidrule{3-7}
     
     & & Batch Learning & $0.208 \pm 0.061$ & $0.084 \pm 0.072$ & $0.062 \pm 0.039$ & $0.013 \pm 0.010$\\
    
     & & FFNN \cite{zhang_learning_2021} & $0.354 \pm 0.124$ & $0.326 \pm 0.125$ & $0.063 \pm 0.036$ & $0.027 \pm 0.012$\\

    \midrule
    \multirow{4}{*}{B} & \multirow{4}{*}{668} & EKF & $1.118 \pm 0.586$ & $0.380 \pm 0.126$ & $0.096 \pm 0.041$ & $0.052 \pm 0.030$\\
    
     & & \textbf{Online Learning} & $0.330 \pm 0.081$ & $0.117 \pm 0.097$ & $0.079 \pm 0.042$ & $0.017 \pm 0.015$\\
    \cmidrule{3-7}
    & & Batch Learning & $0.197 \pm 0.059$ & $0.067 \pm 0.030$ & $0.051 \pm 0.034$ & $0.011 \pm 0.009$\\
    
    & & FFNN \cite{zhang_learning_2021} &  $0.513 \pm 0.223$ & $0.38 \pm 0.181$ & $0.057 \pm 0.034$ & $0.022 \pm 0.011$\\

    \midrule
    \multirow{4}{*}{C} & \multirow{4}{*}{802} & EKF & $0.572 \pm 0.207$ & $0.343 \pm 0.174$ & $0.088 \pm 0.045$ & $0.049 \pm 0.034$\\
    
     & & \textbf{Online Learning} & $0.270 \pm 0.104$ & $0.112 \pm 0.053$ &  $0.081 \pm 0.043$ & $0.033 \pm 0.030$\\
    \cmidrule{3-7}
    & & Batch Learning & $0.178 \pm 0.095$ & $0.086 \pm 0.062$ & $0.047 \pm 0.031$ & $0.019 \pm 0.016$\\
    
    & & FFNN \cite{zhang_learning_2021} & $0.326 \pm 0.102$ & $0.183 \pm 0.058$ & $0.050 \pm 0.031$ & $0.019 \pm 0.013$\\

    \midrule

    \multirow{4}{*}{Overall} & \multirow{4}{*}{2458} & EKF & $0.738 \pm 0.385$ & $0.553 \pm 0.338$ & $0.094 \pm 0.043$ & $0.058 \pm 0.033$\\
    
     &  & \textbf{Online Learning} & $0.292 \pm 0.100$ & $0.115 \pm 0.075$ & $0.076 \pm 0.041$ & $0.023 \pm 0.021$\\
    \cmidrule{3-7}
    & & Batch Learning &  $0.195 \pm 0.076$ & $0.081 \pm 0.062$ & $0.054 \pm 0.036$ & $0.014 \pm 0.012$\\
    
    & & FFNN \cite{zhang_learning_2021} & $0.377 \pm 0.160$ & $0.285 \pm 0.145$ & $0.057 \pm 0.034$ & $0.023 \pm 0.012$\\
    
    \bottomrule
         
    \end{tabular}
    
    \label{tab:table_1}
\end{table*}

\begin{itemize}
    \item \textit{Mean Absolute Trajectory Error (m-ATE)}, which averages the magnitude of the error evaluated between the estimated position and orientation of the robot and its ground truth pose in the same frame. Sometimes, it can lack generalization due to possible error compensations along the trajectory.
    
%    \begin{equation}
%        m-ATE_{\hat{y}} = \dfrac{1}{N}\sum_{n=1}^{N}\left|(\hat{y}_n - y_n)\right|
%    \end{equation}
    
    \item \textit{Segment Error (SE)}, which averages the errors along all the possible segments of a given length s, considering multiple starting points. It is strongly less sensitive to local degradation or compensations than the previous metrics.

%    \begin{equation}
%        SE_{\hat{y}} = \dfrac{1}{N_s}\sum_{n=1}^{N_s}\left|\hat{\textbf{T}}_{n-s}\hat{y}_n - \textbf{T}_{n-s}y_n      \right| %Da finire
%    \end{equation}
\end{itemize}

\subsection{Quantitative Results}\label{subsec:results_q}

The proposed method was tested by training the neural network from scratch using the stream of sensor data in real-time, brought by the ROS 2 topics. The data were first collected in mini-batches of 32 elements. After completion, backpropagation is performed on the model to update all the weights. The data stream is recorded to provide the aforementioned \ref{subsec:paragrafo_exp} training dataset, which was later used to evaluate other methods. The results of the methods are compared to different state-of-the-art solutions, which are i) the same network trained with a traditional batch learning, ii) a feedforward neural network, as in \cite{zhang_learning_2021}, and iii) an Extended Kalman Filter based method, which can be considered one of the most common wheel-inertial odometry estimators.

All the models were evaluated offline using a test set composed of 19 sequences of various lengths, comprised between $60 s$ and $280 s$, which aim to recreate different critical situations for wheel inertial odometry. In particular, the sequences can be separated into three main trajectory types:
\begin{itemize}
    \item \textit{Type A}, comprises round trajectories which do not allow fortunate error compensation during the time. Therefore, they may lead to fast degradation of the estimated pose, and especially of the orientation.
    \item\textit{Type B} comprises an infinite-shaped trajectory. This test allows partial error compensations, but possible unbalanced orientation prediction may lead to fast degradation of the position accuracy. A partial sequence of type B trajectories are shown in Figure \ref{fig:fig_xy}.
    \item \textit{Type C} comprises irregular trajectories, including hard brakings and accelerations, and aims to test the different methods' overall performance.
\end{itemize}

Table \ref{tab:table_1} presents the numeric results of the different tests, considering the proposed model (Online Learning) and the selected benchmarks. All the leaning-based approaches show a significant error reduction compared to the EKF results, which can be considered a baseline for improvement. Considering both the neural network architectures trained offline, the proposed convolutional one achieves an average improvement of 73.5\% on the position $m-ATE_{(x, y)}$ and 85.3\% on the orientation $m-ATE_{\theta}$. In comparison, the FFNN model achieves 49.0\% and 48.4\%, respectively. The Segment Error improves in both cases: the proposed model improves by 42.6\% on the position $SE_{(x, y)}$ and 75.8\% on the orientation $SE_{\theta}$. The FFNN architecture improves by 39.3\% and 60.3\%, respectively.

Compared with the EKF baseline, the online learning model shows almost the same improvement as batch learning. The improvement on the m-ATE equals 60.4\% on the position and 79.2\% on the orientation. The Segment Error also appears to be lower, showing an improvement of 19.1\% on position and 60.3\% on orientation. 
The observed difference between the two training paradigms is an acceptable trade-off between the slight loss of accuracy of the online training compared to the batch training and the possibility of training the model without a pre-collected dataset.

Figure \ref{fig:histograms} reports the histograms of the distribution of the Segment Errors, in position and orientation, respectively, for test scenario B. It emerges how learning-based methods achieve, on average, a smaller error than the EKF method.
Figure \ref{fig:errors} shows the error trend during time related to the trajectory of figure \ref{fig:fig_xy}. It is evident how the batch-trained and online-trained models perform similarly to the other methods.

%%%%%%%%%%%%%%%%%%%%%%%%%%%%%%%%%%%%%%%%%%%%%%%%%%%%%%%%%%%%%%%%%%%%%%%%%%%%%%%%%

\subsection{Latency Evaluation}\label{subsec:further_eval}
Since all the training and inference processes are tested online, firm real-time performance is needed to avoid missing data for training or producing late odometry data. The trained neural network has been converted into a TensorFlow Lite \textit{float32} model, which allows the development of models on edge devices and performs inference on CPU devices. Using the Jackal's Board computer, based on an i3-4330TE @ 2.4 GHz chip, a mean odometry estimation time of 4 ms was achieved on 100 measurements, which is 10\% of the sampling frequency of 25 Hz. The training process on an external PC with 32-GB RAM on a $12^{th}$-generation Intel Core i7 @ 4.7 GHz took an average time of 25 ms per batch, considering 100 measurements.

%%%%%%%%%%%%%%%%%%%%%%%%%%%%%%%%%%%%%%%%%%%%%%%%%%%%%%%%%%%%%%%%%%%%%%%%%%%%%%%%
\section{Conclusions}\label{sec:conclusions}
This paper introduces an online learning approach and an efficient neural network architecture for wheel-inertial odometry estimation in mobile robots from raw sensor data. The online training paradigm does not need a pre-collected dataset and allows fine-tuning the performance of the model over time, adapting to environmental changes. Moreover, the proposed model's reduced dimension allows training and fast inference on a low-resources robotic platform on-the-fly. 

Future works may include developing a collaborative system based on integrating multiple odometry sources with a seamless transition to constantly guarantee accurate localization data to the robot.

\section{Acknowledgement}
This work has been developed with the contribution of Politecnico di Torino Interdepartmental Centre for Service Robotics PIC4SeR\footnote{\url{www.pic4ser.polito.it}}.

\bibliographystyle{unsrt}  
\bibliography{references}  %%% Remove comment to use the external .bib file (using bibtex).


\end{document}